\documentclass[10pt,twocolumn,letterpaper]{article}

\usepackage{cvpr}
\usepackage{times}
\usepackage{epsfig}
\usepackage{graphicx}
\usepackage{amsmath}
\usepackage{amssymb}
\usepackage{booktabs}

\usepackage{amsfonts,bm}
\usepackage{algorithm}
\usepackage{algorithmic}
\usepackage{multirow}
\usepackage{graphicx}
\usepackage{url}
\usepackage{subcaption}

\usepackage{makecell}

\newcommand{\ours}{ActBERT\xspace}

\usepackage[pagebackref=true,breaklinks=true,letterpaper=true,colorlinks,bookmarks=false]{hyperref}

\cvprfinalcopy 

\def\cvprPaperID{7149} 

\ifcvprfinal\fi
\begin{document}

\title{\ours: Learning Global-Local Video-Text Representations}

\renewcommand{\thefootnote}{*}
\author{Linchao Zhu$^{1,2}$ and Yi Yang$^{2*}$ \\
$^{1}$ Baidu Research $^{2}$ ReLER, University of Technology Sydney \\
{\tt\small \{linchao.zhu,yi.yang\}@uts.edu.au}
}

\maketitle
\thispagestyle{empty}

\footnotetext{This work was done when Linchao Zhu visited Baidu Research. Yi Yang is the corresponding author.}

\begin{abstract}

In this paper, we introduce \ours for self-supervised learning of joint video-text representations from unlabeled data. First, we leverage global action information to catalyze mutual interactions between linguistic texts and local regional objects. It uncovers global and local visual clues from paired video sequences and text descriptions for detailed visual and text relation modeling. 
Second, we introduce a TaNgled Transformer block (TNT) to encode three sources of information, \ie, global actions, local regional objects, and linguistic descriptions.
Global-local correspondences are discovered via  judicious  clues extraction from  contextual information. It enforces the joint video-text representation to be aware of fine-grained objects as well as global human intention.
We validate the generalization capability of \ours on downstream video-and-language tasks, \ie, text-video clip retrieval, video captioning, video question answering, action segmentation, and action step localization.
\ours significantly outperforms the state-of-the-art, demonstrating its superiority in video-text representation learning.actbct

\end{abstract}

\section{Introduction}
While supervised learning has been successful in a variety of computer vision tasks~\cite{krizhevsky2012imagenet,he2016deep,c3d,ren2015faster}, self-supervised representation learning from unlabeled data has attracted increasing attention in recent years~\cite{caron2018deep,misra2016shuffle}.
In self-supervised learning, a model is first pre-trained on a large amount of unlabeled data with a surrogate loss.
The fine-tuning process further helps the pre-trained model to be specialized in downstream tasks.
Recently, there has been rapid progress in self-supervised representation learning for texts~\cite{devlin2018bert,yang2019xlnet}, where the Bidirectional Encoder Representations from Transformers (BERT) model~\cite{devlin2018bert} generalizes remarkably to many natural language tasks, \eg, question answering~\cite{alberti2019bert}.

Motivated by BERT's success in self-supervised training, we aim to learn an analogous model for video and text joint modeling.
We exploit video-text relations based on narrated instructional videos, where the aligned texts are detected by off-the-shelf automatic speech recognition (ASR) models.
These instructional videos serve as natural sources for video-text relationship studies. First, they are vastly available and freely accessible on YouTube and other platforms~\cite{miech2019howto100m,sun2019videobert}. Second, the visual frames are aligned with the instructional narrations. The text narrations not only cover the objects in the scene explicitly but identify the salient action in the video clip.

To generalize BERT to video-and-language tasks, Sun~\etal~\cite{sun2019videobert} extended the BERT model by learning from quantized video frame features.
The original BERT takes discrete elements as inputs and predicts the corresponding tokens as the output. In contrast, visual features are distributed representations with real value, while the real-value features cannot be directly categorized into discrete labels for ``visual token'' prediction.  Sun~\etal~\cite{sun2019videobert} discretized visual features into visual words via clustering. These visual tokens can be directly passed to the original BERT model. 
However, detailed local information, \eg, interacting objects, human actions would be possibly lost during clustering. It prevents the model from uncovering fine-grained relations between video and text.
In this paper, we propose \ours to learn a joint video-text representation that uncovers global and local visual clues from paired video sequences and text descriptions.
Both the global and the local visual signals interact with the semantic stream mutually. \ours leverages profound contextual information and exploits fine-grained relations for video-text joint modeling.

First, \ours incorporates global actions, local regional objects and text descriptions in a joint framework.
Actions, \eg, ``cut'', ``rotate'', ``slice'', are essential to various video-related downstream tasks. The recognition of human actions can demonstrate the model's capacity in motion understanding and complex human intention reasoning. It could be beneficial to explicitly model human actions during model pre-training. 
Long-term action sequences furthermore offer temporal dependencies about an instructional task.
Though action clues are important, they are largely ignored in previous self-supervised video-text training~\cite{sun2019videobert,miech2019howto100m}, where actions are treated identically to objects.
To model human actions, we first extract verbs from the text descriptions and construct an action classification dataset from the original dataset. Then, a 3D convolution network is trained to predict the action labels. The features from the optimized network are used as the action embedding.
In this way, clip-level actions are represented, and the corresponding action label is inserted.
Besides global action information, we incorporate local regional information to provide fine-grained visual cues~\cite{lu2019vilbert,tan2019lxmert,su2019vl,li2019unicoder,chen2019uniter}. Object regions provide detailed visual clues about the whole scene, including the regional object feature, the position of the object. The language model can benefit from the regional information for better language-and-visual alignment.

Second, we introduce a TaNgled Transformer block (TNT) to encode features from three sources, \ie, global actions, local regional objects, and linguistic tokens.
Previous studies \cite{lu2019vilbert,tan2019lxmert} consider two modalities when designing the new transformer layers, \ie, fine-grained object information from image and natural language. Lu \etal~\cite{lu2019vilbert} introduced a co-attentional transformer layer, where the key-value pairs from one modality are passed to the other modality's attention block to act as the new key-value pairs.
However, in our scenario, there are three sources of inputs. 
The two sources, \ie, local regional features and linguistic texts, offer detailed descriptions of the occurring event in the clip. The other global action feature provides the human intention in time-series as well as a straightforward clue for contextual inferring.
We design a new tangled transformer block for cross-modality feature learning from three sources.
To enhance the interactions between two visual cues and linguistic features, we use a separate transformer block~\cite{vaswani2017attention} to encode each modality.
The mutual cross-modal communication is later enhanced with two additional multi-head attention blocks.
The action feature catalyzes mutual interactions. With the guidance from the action features,
we inject visual information to the linguistic transformer, and incorporate linguistic information to the visual transformers. The tangled transformer dynamically selects judicious cues its context to facilitate the target prediction.

Furthermore, we design four surrogate tasks to train \ours, \ie, masked language modeling with global and local visual cues, masked action classification, masked object classification and cross-modal matching. The pre-trained \ours is transferred to five video-related downstream tasks, \ie, video captioning, action segmentation, text-video clip retrieval, action step localization, and video question answering.
We quantitatively show \ours achieves the state-of-the-art performance with a clear margin.

\section{Related Work}
\noindent\textbf{Video and language.}
There are many existing video-and-language tasks to evaluate the model's capacities in joint video-text representation learning, e.g., video question answering \cite{tapaswi2016movieqa,jang2017tgif,lei2018tvqa,zhu2017uncovering}, video captioning~\cite{yao2015describing,zhou2018end}, text-video retrieval~\cite{yu2018joint,wang2019learning,miech2018learning}, video grounding~\cite{zhou2019grounded}.
In video and language modeling, it can be difficult to learn relations between ordered video frames and their corresponding descriptions, where video temporal information and the interactions between multiple objects spatio-temporally requires to be incorporated. The dominant approach for multi-modal modeling is to leverage Recurrent Neural Networks (RNNs) and their variants, \eg, Long Short-Term Memory (LSTM) and Gated Recurrent Unit (GRU), to model sequence relations, \eg, \cite{pan2016jointly,zhu2017bidirectional}. Zhou~\etal\cite{zhou2018end} leveraged masked transformers in both the encoder and the decoder for dense video captioning.
Most of these works are conducted on well-annotated datasets where the descriptions are manually generated, requiring considerable human interference. 
There are other works to learn video representations from limited annotated data \cite{zhu2018compound}.
The video data is a natural source to learn cross-modal representations. The text descriptions are automatically generated by off-the-shelf automatic speech recognition (ASR) models. This is more scalable and general to the model's deployment in real-world applications.
In this paper, we focus on learning joint video-text representation in a self-supervised way. 

\noindent\textbf{Cross-modal pre-training.}
In the past year, many works extended BERT to model cross-modal data~\cite{lu2019vilbert,su2019vl,tan2019lxmert,chen2019uniter,li2019unicoder,sun2019videobert}. The recent BERT model for video-text modeling~\cite{sun2019videobert} introduces visual words for video frames encoding, where local regional information is largely ignored.
The synchronized video-audio signal is also a good test-bed for cross-modal representation learning~\cite{arandjelovic2018objects,korbar2018cooperative}. However, they leveraged low-level audio signals and only considered  the synchronization nature of video data. In this work, we focus on video-text joint representation learning. Our \ours leverages multi-source information and achieves remarkable performance in many downstream  video-text tasks.

\noindent\textbf{Instructional videos.}
Learning from instructional videos is challenging  due to its data complexity across various tasks~\cite{damen2018scaling,alayrac2016unsupervised,zhou2018towards,miech2019howto100m}.
These videos are collected from many domains, \eg, cooking, sports, gardening.
Many works also regard the transcriptions generated from instructional videos as a source of supervision~\cite{alayrac2016unsupervised,zhou2018towards,miech2019howto100m}. However, we employ \ours to explicitly model human actions, local regions in a unified framework. We improve \cite{miech2019howto100m} with more specific relation modeling between videos and their description. We quantitatively demonstrated that  \ours is more suitable for unsupervised video-text modeling.

\section{Model Architecture}
\subsection{Preliminary}
\label{sec:preliminary}
We first illustrate the original BERT~\cite{devlin2018bert} model.
BERT~\cite{devlin2018bert} pre-trains a language model on large corpora in an unsupervised way. The pre-trained model is found to be flexible and beneficial to a variety of downstream tasks, \eg, question answering~\cite{alberti2019bert}.

In BERT \cite{devlin2018bert}, the input entities are processed by a multi-layer bidirectional transformer~\cite{vaswani2017attention}. The embeddings of each input are processed with stacked self-attention layers to aggregate contextual features. The attention weights are adaptively generated. The output features contain contextual information about the original input sequence.
In self-attention, the generated features are irrelevant to input sequence order, and it enables the output representation to be permutation-invariant.  The output representation is not affected when the input sequence is shuffled. A position embedding is commonly applied to each input entity for the incorporation of sequential order clues.

In the original BERT, Devlin \etal introduced two tasks for pre-training.
In the task of masked language modeling (MLM), a portion of input words are randomly masked out. These masked-out words are replaced by a special token ``[MASK]''.
The task is to predict the masked words based on the observations from the contextual contents.
The contextual contents are unmasked elements that provide useful relevant cues for the prediction of the masked word.

The other task, \ie, Next Sentence Prediction (NSP), models order information between two sentences. Two sentences are sampled from a document, and NSP aims to identify if the second sentence is adjacent to the first sentence with the correct order. The two sentences are concatenated via a token ``[SEP]'', so that the models can be aware of the inputs being separated sentences. The prediction is made upon the output features of the first token ``[CLS]''. This is a binary classification problem, and a simple sigmoid classifier is used. A prediction of ``$1$'' indicates the sentences are consecutive, and the second sentence is right after the first sentence.

\subsection{\ours}
\subsubsection{Input Embeddings}
There are four types of input elements in \ours. They are actions, image regions, linguistic descriptions and special tokens. Special tokens are used to distinguish different inputs.

Each input sequence starts with a special token ``[CLS]'' and ends with another token ``[SEP]''. 
We put the linguistic descriptions after ``[CLS]''. There are the action inputs followed by local regional features.
We denote the action features as $a_1, \ldots, a_L$, the frame region features as $r_1, \ldots, r_M$. The sequential text descriptions is denoted as $w_1, \ldots, w_N$.
The whole sequence is denoted as $\{\text{[CLS]}, w_1, \ldots, w_N, \text{[SEP]}, a_1, \ldots, a_L, \text{[SEP]},$ $r_1, \ldots, r_M, \text{[SEP]}\}$. ``[SEP]'' is also inserted between different sentences. We also insert ``[SEP]'' between regions that are from different clips, which can help the model to identify the clip boundaries.
For each input step, the final embedding feature consists of four different embeddings.
The embeddings are position embedding, segment embedding, token embedding, visual feature embedding. We added a few new tokens to distinguish action features and regional object features. The visual embedding is introduced to extract visual and action information.
These embeddings are added to be the final feature of \ours. We explain them in detail as follows.

\noindent\textbf{Position embedding.}
Following~\cite{devlin2018bert}, we incorporate a learnable position embedding to every input in the sequence. Since self-attention does not consider order information, position encoding offers a flexible way to embed a sequence when the sequence order matters.
For the actions in different clips, the position embeddings are different as the video clips are ordered. For the regions extracted from the same frame, we use the same position embedding.
To distinguish regions from the same frame, we consider spatial position embedding for different spatial positions. The details will be described in ``Visual (action) embedding''.

\noindent\textbf{Segment embedding.}
We consider multiple video clips for long-term video context modeling.
Each video clip or video segment has a corresponding segment embedding.
The elements, \ie, action inputs, regional object inputs, linguistic descriptions, have the same segment embedding in the same video clip.

\noindent\textbf{Token embedding.}
Each word is embedded with WordPiece embeddings~\cite{wu2016google} with a 30,000 vocabulary.
In addition to the special tokens mentioned above (``[CLS]'', ``[MASK]'', ``[SEP]''), we introduce ``[ACT]'' and ``[REGION]'' to represent the action features and the region features extracted from video frames, respectively. Note that all action inputs have the identical token embedding, which reveals the modality of the inputs.

\noindent\textbf{Visual (action) embedding.}
We now explain the visual (action) embedding in details. We first illustrate the procedure to obtain the action embedding. 
For each video clip, we extract verbs from its corresponding descriptions.
For simplicity, we remove clips that do not have any verbs. We then build a vocabulary from all the extracted verbs. After verb vocabulary construction, each video clip has one or multiple category labels. We train a 3D convolutional neural network on this constructed dataset. The inputs to the 3D network is a tensor that contains an additional temporal dimension. We leverage a softmax classifier on top of the convolutional neural network. For clips with multiple labels, we normalize the one-hot label with $\ell_1$-norm, where the scores for all labels are summed to be $1$. After the model is trained, we extract the features after global average pooling as the \textbf{action features}. This feature can well represent the actions that occurred in the video clip.

To obtain regional object features, we extract bounding boxes and the corresponding
visual features from a pre-trained object detection network.
Similar to Lu~\etal~\cite{lu2019vilbert}, we utilized pre-trained Faster R-CNN network \cite{ren2015faster} to extract the categorical distribution under the COCO vocabulary \cite{lin2014microsoft}. The image region features offer detailed visual information for visual and text relation modeling. 
For each region, the visual feature embeddings are the feature vectors before the output layer in the pre-trained network.
Following~\cite{lu2019vilbert}, we incorporate spatial position embeddings to represent region locations with a 5-D vector. This vector consists of four box coordinates and the fraction of the region area. Specifically, we denote the vector as $(\frac{x_1}{W}, \frac{y_1}{H}, \frac{x_2}{W}, \frac{y_2}{H}, \frac{(x_2-x_1)*(y_2-y_1)}{W*H})$, where $W$ is the frame width, $H$ is the frame height, and $(x_1, y_1)$ and $(x_2, y_2)$ are the top-left and bottom-right coordinates, respectively.

This vector is then embedded to match the dimension of the visual feature. The final regional object feature is the summation of the spatial position embedding and the object detection feature.

\subsubsection{Tangled Transformer}
We design a TaNgled Transformer (TNT) to better encode three sources of information, \ie, action features, regional object features and linguistic features.

Instead of using only one transformer that treats the visual and text features equally, our tangled transformer consists of three transformers. The three transformers take three sources of features, respectively. To enhance the interactions between visual and linguistic features, we propose to inject visual information to the linguistic transformer and incorporate linguistic information to the visual transformers. With cross-modal interactions, the tangled transformer can dynamically select judicious cues for target prediction.

\begin{figure}[t]
\centering
\includegraphics[width=0.95\linewidth]{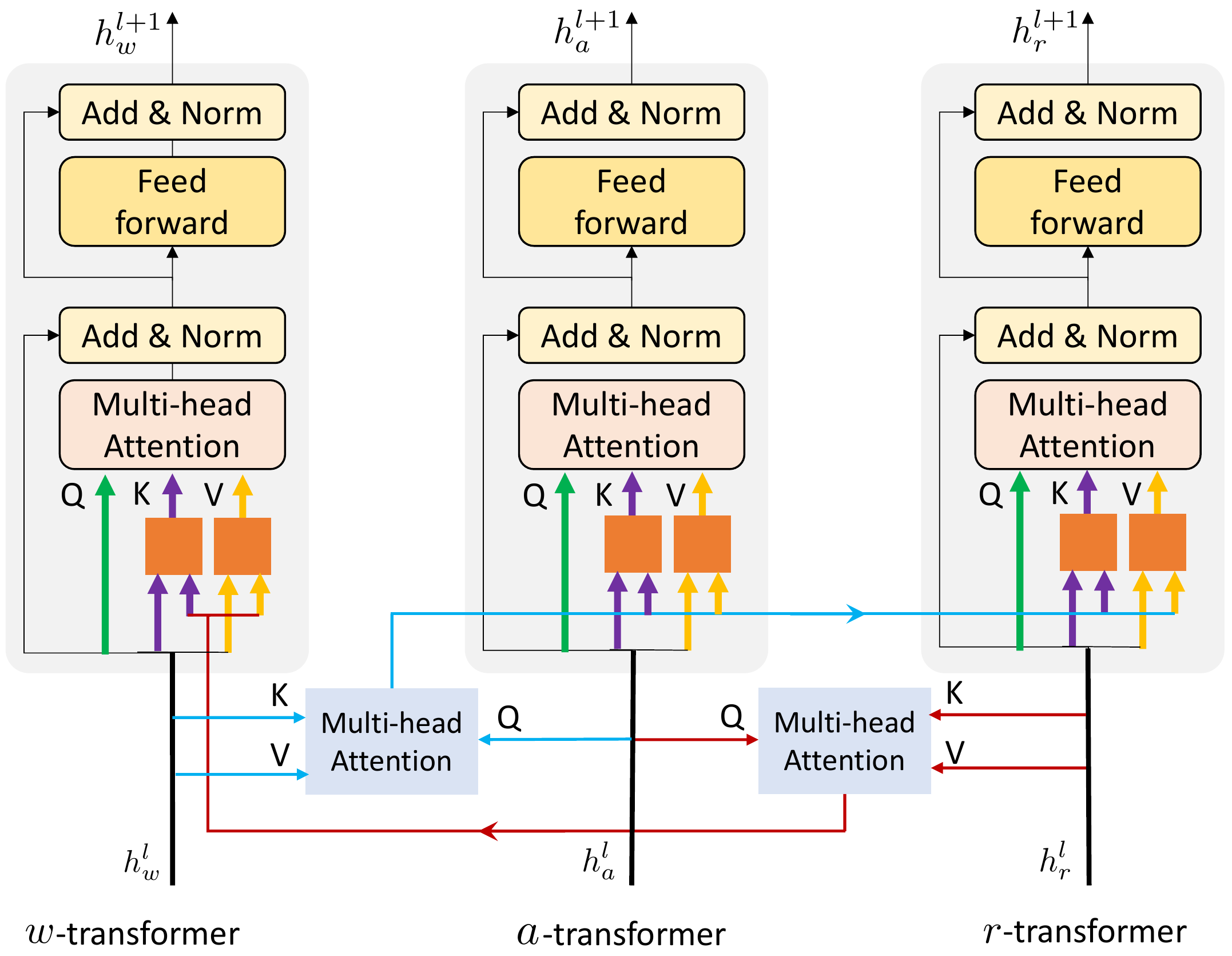}
    \caption{\textbf{Our tangled transformer} takes three sources of information as inputs, which enhances the interactions between linguistic features and visual features.}
    \label{fig:our_transformer}
\end{figure}

We denote the intermediate representations at transformer block $l$ as $h^l=\{(h_{w_0}^l, $ $\ldots, h_{w_N}^l), (h_{a_0}^l, \ldots, h_{a_L}^l), (h_{r_0}^l, \ldots, h_{r_M}^l)\}$. For simplicity, we denote $h_{w}^l=\{h_{w_0}^l, \ldots, h_{w_N}^l\}$, $h_{a}^l=\{h_{a_0}^l, \ldots, h_{a_L}^l)\}$, and
$h_{r}^l=\{h_{r_0}^l, \ldots, h_{r_M}^l)\}$, which are processed by $w$-transfomer, $a$-transformer, and $r$-transformer, respectively (Figure~\ref{fig:our_transformer}).
Besides the standard multi-head attention encoding features from the same modality, we leverage the other two multi-head attention blocks to enhance  mutual interactions between the transformer blocks.
Specifically, we utilize $h_a^l$ to catalyze mutual interactions. We denote the multi-head attention as $output=Multihead(Q, K, V)$, where $Q$ is the query, $K$ is the key, $V$ is the value. The details of multi-head attention can be found in \cite{vaswani2017attention}.
We use $h_a^l$ as a query to attend judicious cues from $h_w^l$ and $h_r^l$:
\begin{align}
    c_w &= Multihead(W_{q}^1 h_a^l, W_{k}^w h_w^l, W_{v}^w h_w^l), \\
    c_r &= Multihead(W_{q}^2 h_a^l, W_{k}^r h_r^l, W_{v}^r h_r^l),
\end{align}
where $W^{*}_{*}$ are learnable weights. $c_w$ is the blended feature from linguistic representations, while $c_r$ is the guided feature from regional object representation. We then generate a new key-value pair from $c_w$ using a linear layer. This generated key-value pair is stacked with the key-value pairs from the original $a$-transformer and $r$-transformer. Similarly, we generate a new key-value pair from $c_r$, which is stacked with key-value pair in $w$-transformer.
With this form tangled transformer, visual and linguistic features are further associated.

Note that our tangled transformer is different from the co-attentional transformer block in~\cite{lu2019vilbert} in several ways.
First, the co-attentional transformer block simply passes the keys and values
from one modality to the other modality's attention block, without further pre-processing. Second, \cite{lu2019vilbert} treats the two modalities equally, while our tangled block utilizes a global cue to guide the selection of local hints from linguistic and visual features. Third, the keys and values from different modalities replace the origin key-values in \cite{lu2019vilbert}, while our tangled transformer stacks the key-value with the original one. In this way, both the linguistic and visual features are incorporated during transformer encoding.

\subsubsection{\ours Training}

\begin{figure*}[t]
\centering
\includegraphics[width=0.8\linewidth]{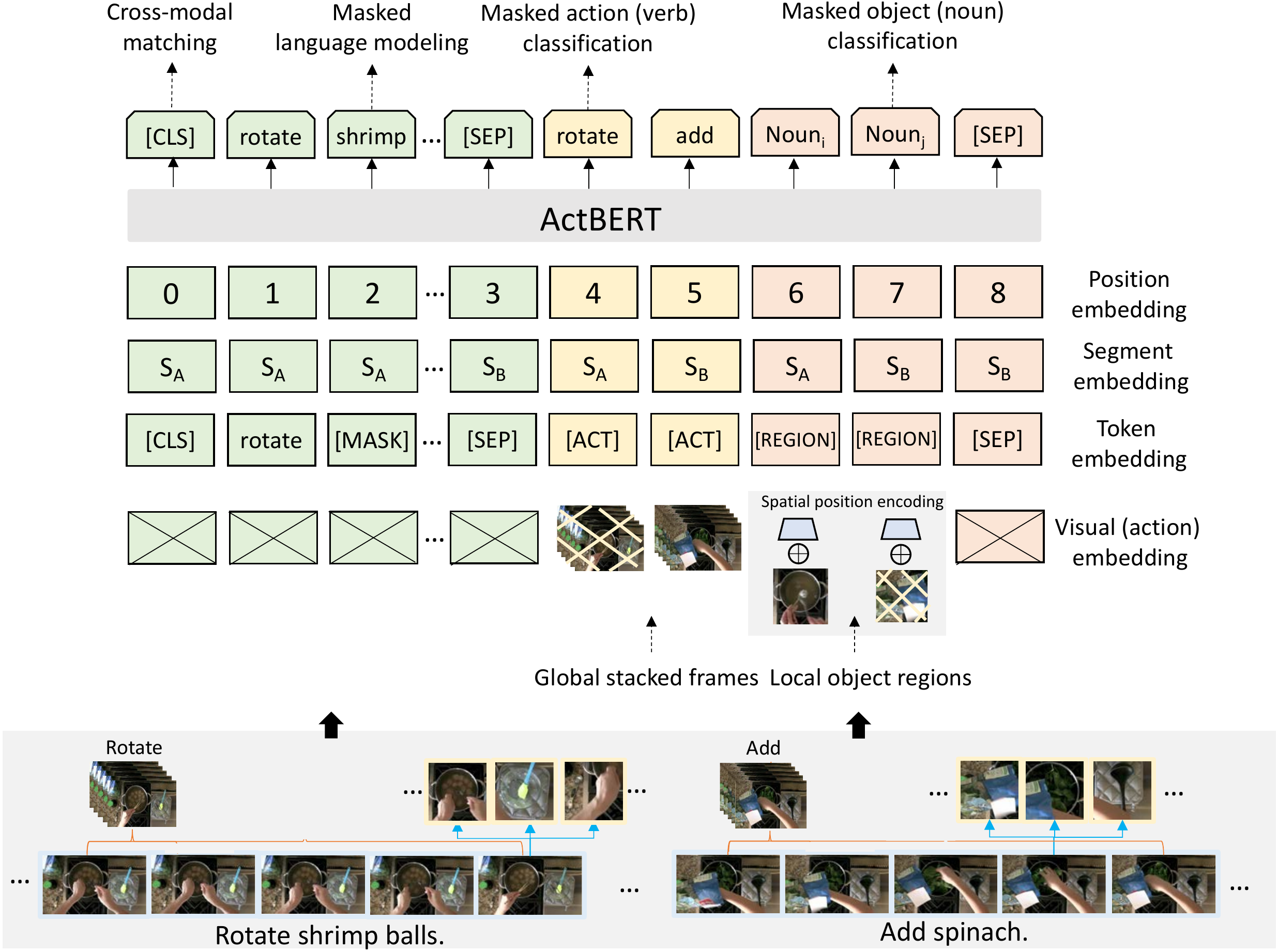}
    \caption{\textbf{Our \ours framework}. We incorporate three sources of information during pre-training, \ie, global actions, local regional objects, and text descriptions.
    The yellow grid indicates that the action or the region object is masked out.
    }
    \label{fig:framework}
\end{figure*}

We introduce four tasks for \ours pre-training. Our framework is presented in Figure \ref{fig:framework}.
We naturally extend the Masked Language Modeling in our cross-modal setting. There are some existing extensions for image and language pre-training~\cite{lu2019vilbert,sun2019videobert}, and video and language pre-training~\cite{sun2019videobert}. Compared to \cite{sun2019videobert}, we explicitly model actions and regional information in a unified framework. 

\noindent\textbf{Masked Language Modeling with Global and Local Visual Cues.}
We extend the Masked Language Modeling (MLM) task in BERT to our setting. We leverage visual cues from local regional objects and global actions to uncover the relationships between visual and linguistic entities. As described in Section~\ref{sec:preliminary}, each word in the input sentence is randomly masked with a fixed probability.
The task forces the model to learn from contextual descriptions, and at the same time, extract relevant visual features to facilitate prediction.
When a verb word is masked out, the model should exploit the action features for a more accurate prediction. When a description of an object is masked out, local regional features can provide more contextual information. Thus, the strong model needs to align visual and linguistic inputs locally and globally.
The output feature is then appended with a softmax classifier over the whole linguistic vocabulary. 

\noindent\textbf{Masked Action Classification.}
Similarly, in Masked Action Classification, the action features are masked out. The task is to predict the masked action label based on linguistic features and object features.
Explicit action prediction can be beneficial in two perspectives. First, action sequential cues can be exploited in the long-term. For example, for a video with action sequences of ``get into'', ``rotate'', ``add'', this task can better exploit the temporal order information regarding performing this instructional assignment.
Second, the regional objects and linguistic texts are leveraged for better cross-modality modeling. Note that in Masked Action Classification, the goal is to predict the categorical label of the masked-out action feature.
This task can enhance the action recognition capability of the pre-trained model, which can be further generalized to many downstream tasks, \eg, video question answering.

\noindent\textbf{Masked Object Classification.}
In Masked Object Classification, the regional object features are randomly masked out. We follow~\cite{lu2019vilbert} to predict a distribution over fixed vocabulary for the masked-out image region. 
The target distribution of the masked-out region is calculated as the softmax activation that is extracted by forwarding the region to the same pre-trained detection model in the feature extraction stage. The KL divergence between the two distributions is minimized.

\noindent\textbf{Cross-modal matching.}
Similar to the Next Sentence Prediction (NSP) task, we apply a linear layer on top of the output of the first token ``[CLS]''.
It is followed by a sigmoid classifier, indicating the relevance score of the linguistic sentences and the visual features. If the score is high, it shows that the text well-describes the video clips.
The model is optimized via a binary cross-entropy
loss. To train this cross-modal matching task, we sample negative video-text pairs from the unlabeled dataset. We follow \cite{miech2019howto100m} for sampling positive pairs and negative pairs.

\section{Experiments}
\label{sec:expr}
In this section, we evaluate \ours in multiple downstream video-and-language tasks.
We quantitatively evaluate the generalization capability of \ours on five challenging tasks, \ie, text-video clip retrieval, video captioning, video question answering, action segmentation, and action step localization.

\subsection{\ours implementation details}
\paragraph{HowTo100M.} We pre-train \ours on the HowTo100M dataset~\cite{miech2019howto100m}.
The HowTo100M dataset is constructed by querying YouTube API. The top 200 search results are kept.
This dataset covers a total of 23,611 tasks, \eg, maintenance and repair, animal rescue, food preparation. This dataset is biased towards actions, where the verbs like ``go'', ``make'', ``come'' being the most frequent. The nouns are also distributed in a long-tailed way, where objects like ``water'', ``cup'' are ranked top.
Each video has a corresponding narration that is extracted from video subtitles.
As the association between video clips and texts are not manually annotated, the video-text connection can sometimes be weak. There are cases of noisy correspondences, where the actors sometimes talk about unrelated things. Though noisy, we found pre-training on HowTo100M can still significantly improve the performance of downstream tasks.

\noindent\textbf{Pre-training details.} 
To construct video-text inputs for \ours pre-training, we sample video clips from the HowTo100M dataset.
Instead of only using one clip for video-text joint training, we leverage multiple adjacent clips to cover a longer context. This enables \ours to model relations in different segments. We sample 10 adjacent video clips, and the temporal-aligned linguistic tokens are extracted to form a video-text pair. 

To obtain the local regional features, we use Faster R-CNN pre-trained on the Visual Genome~\cite{krishna2017visual} dataset following \cite{lu2019vilbert}. The backbone is ResNet-101~\cite{he2016deep}.
We use the frame rate of 1 FPS to extract the regional features.
Each region feature is RoI-pooled from the convolutional feature from that region.
We set the detection confidence threshold as 0.4, and each frame contains at most five boxes.
Transformer and co-attentional transformer blocks in the visual stream have hidden state size of 1024
and 8 attention heads.

To obtain the action features, we first construct an action classification dataset. We sample frames at 8 FPS. For each clip, we extract the verb from its text descriptions. Then, we train a ResNet-3D~\cite{tran2018closer} network with a softmax classification loss.
We initialized the weights of the ResNet-3D model from a pre-trained model on Kinetics~\cite{kay2017kinetics}.
The Kinetics dataset covers 400 actions
from YouTube videos. The 3D convolutional network converges faster using when it is pre-trained on Kinetics.
The input clip length to ResNet-3D is 32. The clip covers a 4-second video duration. The spatial shape of the input frame is 224$\times$224.
The initial learning rate is set to 0.001. The batch size is 16. We decay the learning rate by 0.1 at iteration 100,000, and the total number of training iterations is 1,000,000. We keep other training settings unchanged following~\cite{tran2018closer}. During feature extraction, we sample the central clip, and each frame is central cropped. We use the feature after global average pooling as the clip representation.

During \ours pre-training, 15\% of input features are randomly masked out. \ours has 12 layers of transformer blocks. Each transformer block has a hidden unit size of 768.
We initialize the linguistic transformer with the BERT model pre-trained on the BookCorpus \cite{zhu2015aligning} and English Wikipedia.
The other two transformers are randomly initialized.
The network is optimized by Adam optimizer. We set the learning rate to be $10^{-5}$.
We trained the model for five epochs due to the large-scale data. 

\subsection{Results on video-and-text tasks}
We evaluate \ours on five downstream tasks, \ie, action step localization, action segmentation, text-video clip retrieval, video captioning, and video question answering.
We evaluate the five tasks on  CrossTask~\cite{zhukov2019cross}, COIN~\cite{tang2019coin},
YouCook2~\cite{zhou2018towards}, MSR-VTT~\cite{xu2016msr} and LSMDC \cite{lsmdc}.
Videos from the test sets of these datasets are removed during pre-training on HowTo100M.

\subsubsection{Datasets}

\paragraph{CrossTask:}
We evaluate action step localization on the CrossTask~\cite{zhukov2019cross} dataset.
CrossTask~\cite{zhukov2019cross} contains 83 tasks and 4.7k videos related to cooking, car maintenance,
crafting, etc. 
We use the recall metric described in~\cite{zhukov2019cross}, which is defined by the number of step assignments that fall into the ground-truth interval, divided by the total number of steps in the video.
\textbf{COIN:}
We evaluate the action segmentation task on the recent COIN~\cite{tang2019coin} dataset.
COIN~\cite{tang2019coin} contains 180 tasks and 11,827 videos. This dataset consists of 46,354 annotated segments.
The videos are collected from YouTube.
\textbf{YouCook2:} 
We evaluate text-video clip retrieval and video captioning on YouCook2.
YouCook2 is a cooking video dataset collected from YouTube, covering a large variety of cooking styles, methods, ingredients and cookwares~\cite{zhou2018towards}.
In YouCook2, there are 89 types of recipes and totally 14k clips described with linguistic texts. 
Following~\cite{miech2019howto100m}, we evaluate the text-video clip retrieval task on the validation clips of YouCook2.
\textbf{MSR-VTT:} 
We evaluate text-video clip retrieval and video question answering on MSR-VTT. The MSR-VTT dataset \cite{xu2016msr} is a general video dataset collected from YouTube with text descriptions.  
For the video question answering task, we evaluate the multiple-choice VideoQA following~\cite{yu2018joint}. There are 2,990 questions in total for testing. Each test video is associated with a ground-truth caption, a correct answer, and four mismatched descriptions.
For text-video clip retrieval, following~\cite{yu2018joint}, we use 1,000 pairs text-video for evaluation.
\textbf{LSMDC}: We evaluate fill-in-the-blank video question answering on LSMDC \cite{lsmdc}.
This task is to predict a single answer given a video clip and a sentence with a blank in it.
In LSMDC fill-in-the-blank, there are 296,960 training question-answer pairs, 21,689 validation pairs, and 30,349 testing pairs on public test set. The accuracy is reported on the public test set.

\subsubsection{Video captioning}
We compare our \ours to VideoBERT \cite{sun2019videobert} on the video captioning task.
We take the pre-trained action transformer as the video encoder. We follow the setup from \cite{zhou2018end} that takes the video clips from YouCook2 \cite{zhou2018towards} as input, and a transformer decoder is used to decode videos to captions. We do not use the regional object transformer to fairly compare to \cite{sun2019videobert}. Similar to \cite{sun2019videobert}, we cross-validate the  hyper-parameters on the training set.
We report the standard evaluation metrics for captioning, \ie, BLEU, METEOR, and ROUGE, on the validation set.
The model is optimized by Adam optimizer for 40k iterations. We set the initial learning rate to $1.0\times10^{-3}$, and the batch size is 128.
The results are shown in Table~\ref{tab:youcook_captioning}.
We outperform VideoBERT \cite{sun2019videobert} across all metrics, achieving a 1.36 improvement on METEOR. It demonstrates that our pre-trained transformer learns a better video representation. It also indicates the effectiveness of \ours in modeling video sequences by considering both global and local video cues. Our transformer generalizes better in video captioning. 

\begin{table}
\begin{center}
        \resizebox{\linewidth}{!}{
\begin{tabular}{ lccccc }
\toprule
Method & BLEU-3 & BLEU-4 & METEOR & ROUGE-L & CIDEr  \\
\midrule
Zhou \etal~\cite{zhou2018end} & 7.53 & 3.84 & 11.55 & 27.44 & 0.38 \\
S3D~\cite{xie2018rethinking} & 6.12 & 3.24 & 9.52 & 26.09 & 0.31 \\
VideoBert \cite{sun2019videobert} & 6.80 & 4.04 & 11.01 & 27.50 & 0.49\\
VideoBert + S3D \cite{sun2019videobert}& {7.59} & {4.33} & {11.94} & {28.80} & {0.55}\\
\midrule
\ours & \textbf{8.66} & \textbf{5.41} & \textbf{13.30} & \textbf{30.56} & \textbf{0.65} \\
\bottomrule
\end{tabular}
}
\end{center}
\caption{\textbf{Video captioning} results on YouCook2. We outperform VideoBERT \cite{sun2019videobert} across all the metrics.
}
\label{tab:youcook_captioning}
\end{table}

\begin{table}[tb]
\small
\centering
\begin{tabular}{l  c }
\toprule
Method & Frame Accuracy (\%) \\
\toprule
NN-Viterbi~\cite{richard2018neuralnetwork} & 21.17\\
VGG~\cite{simonyan2014very} & 25.79 \\
TCFPN-ISBA~\cite{ding2018weakly} & 34.30\\
\midrule 
\ours w/o region cues & 52.10 \\
\ours & \textbf{56.95} \\
\bottomrule
\end{tabular}
\caption{\textbf{Action segmentation} results on COIN.} \label{tab:action_segmentation}
\end{table}

\begin{table*}[t]
\resizebox{\textwidth}{!}{
\begin{tabular}{lc@{~~~~}c@{~~}c@{~~}c@{~~}c@{~~}c@{~~}c@{~~}c@{~~}c@{~~}c@{~~}c@{~~}c@{~~}c@{~~}c@{~~}c@{~~}c@{~~}c@{~~}c@{~~}|c} \toprule
    & \rotatebox{90}{\small Make} \rotatebox{90}{\small Kimchi Rice}  
    & \rotatebox{90}{\small Pickle} \rotatebox{90}{\small Cucumber}  
    & \rotatebox{90}{\small Make Banana} \rotatebox{90}{\small Ice Cream}  
    & \rotatebox{90}{\small Grill} \rotatebox{90}{\small Steak}  
    & \rotatebox{90}{\small Jack Up } \rotatebox{90}{\small Car}  
    & \rotatebox{90}{\small Make } \rotatebox{90}{\small Jello Shots}  
    & \rotatebox{90}{\small Change } \rotatebox{90}{\small Tire}  
    & \rotatebox{90}{\small Make } \rotatebox{90}{\small Lemonade}  
    & \rotatebox{90}{\small Add Oil } \rotatebox{90}{\small to Car}  
    & \rotatebox{90}{\small Make } \rotatebox{90}{\small Latte}  
    & \rotatebox{90}{\small Build } \rotatebox{90}{\small Shelves}  
    & \rotatebox{90}{\small Make } \rotatebox{90}{\small Taco Salad}  
    & \rotatebox{90}{\small Make } \rotatebox{90}{\small French Toast}  
    & \rotatebox{90}{\small Make } \rotatebox{90}{\small Irish Coffee}  
    & \rotatebox{90}{\small Make } \rotatebox{90}{\small Strawberry Cake}  
    & \rotatebox{90}{\small Make } \rotatebox{90}{\small Pancakes}  
    & \rotatebox{90}{\small Make } \rotatebox{90}{\small Meringue}  
    & \rotatebox{90}{\small Make } \rotatebox{90}{\small Fish Curry}  
    & \rotatebox{90}{\small Average }
\\ \midrule
Alayrac \etal \cite{alayrac2016unsupervised} & 15.6 & 10.6          & 7.5           & 14.2          & 9.3           & 11.8          & 17.3          & 13.1          & 6.4 & 12.9          & 27.2          & 9.2           & 15.7          & 8.6           & 16.3          & 13.0          & 23.2          & 7.4           & 13.3 \\
Zhukov \etal \cite{zhukov2019cross}                         & 13.3          & 18.0          & 23.4 & 23.1 & 16.9          & 16.5 & 30.7 & 21.6 & 4.6          & 19.5 & 35.3        & 10.0 & 32.3 & 13.8 & 29.5 & 37.6 & {43.0} & 13.3 & 22.4 \\ 
Supervised \cite{zhukov2019cross}                           & 19.1          & 25.3          & 38.0          & 37.5          & 25.7          & 28.2          & \textbf{54.3}          & 25.8          & 18.3         & 31.2          & 47.7          & 12.0          & 39.5          & 23.4          & 30.9          & 41.1          & \textbf{53.4}          & 17.3          & 31.6 \\ 
TVJE~\cite{miech2019howto100m}                    & {33.5}           & {27.1}         & {36.6}           & {37.9}           & {24.1}           & {35.6} & {32.7}           & {35.1}           & {30.7}         & {28.5}           & {43.2}         & {19.8}           & {34.7}           & {33.6}           & {40.4}           & {41.6}          & 41.9           & {27.4}           & {33.6} \\
\midrule 
\ours w/o region cues & {37.4} &    {29.5} &    {39.0} &    {42.2} &    {29.8} &    {37.5} &    {35.5} &    {37.8} &    {33.2} &    {32.8} &    {48.4} &    {25.2} &    {37.4} &    {35.6} &    {42.4} &    {47.0} &    {46.1} &    {30.4} & {37.1} \\
\ours & \textbf{41.8} &    \textbf{33.6} &    \textbf{42.7} &    \textbf{46.8} &    \textbf{33.4} &    \textbf{43.0} &    \textbf{40.8} &    \textbf{41.8} &    \textbf{38.3} &    \textbf{37.4} &    \textbf{52.5} &    \textbf{30.1} &    \textbf{41.2} &    \textbf{40.4} &    \textbf{46.1} &    \textbf{51.0} &    \textbf{49.7} &    \textbf{35.1} & \textbf{41.4} \\
\bottomrule
\end{tabular}
}
\caption{\textbf{Action step localization} results on CrossTask~\cite{zhukov2019cross}.}
\label{table:action_step_localization}
\end{table*}

\begin{table}[t]
    \centering  
        \resizebox{\linewidth}{!}{
      \begin{tabular}{lccccc}
      \toprule
      Method & Dataset  & R@1 & R@5 & R@10  & Median R  \\
            \midrule
      HGLMM~\cite{klein2015associating} & YouCook2 & 4.6 & 14.3 & 21.6 & 75   \\
      TVJE~\cite{miech2019howto100m} & YouCook2 & 4.2 & 13.7 & 21.5 & 65 \\
      TVJE +FT~\cite{miech2019howto100m} & YouCook2 & {8.2} & {24.5} & {35.3} & {24} \\
      \midrule
      \ours & YouCook2 & {9.6} & {26.7} & {38.0} & {19} \\
      \midrule
      \midrule
      C+LSTM+SA~\cite{torabi2016learning}     & MSR-VTT  & 4.2 & 12.9 & 19.9 & 55 \\
      VSE-LSTM~\cite{kiros2014unifying} & MSR-VTT  & 3.8 & 12.7 & 17.1 & 66 \\ 
      SNUVL~\cite{yu2016video} & MSR-VTT & 3.5 & 15.9 & 23.8 & 44\\
      Kaufman \etal~\cite{kaufman2017temporal}  & MSR-VTT  & 4.7 & 16.6 & 24.1 & 41\\
      CT-SAN~\cite{yu2017end}   & MSR-VTT  & 4.4 & 16.6 & 22.3 & 35\\
      JSFusion~\cite{yu2018joint} & MSR-VTT & 10.2 & 31.2 & 43.2 & 13  \\
      TVJE \cite{miech2019howto100m} & MSR-VTT & 7.5 & 21.2 & 29.6 & 38 \\
      TVJE +FT \cite{miech2019howto100m} & MSR-VTT & {14.9} & {40.2} & {52.8} & {9} \\
      
            \midrule
      \ours & MSR-VTT & 8.6 & 23.4 & 33.1 & 36 \\
      \ours +FT        & MSR-VTT & 16.3     &              42.8         &      56.9             &     10 \\
      \bottomrule
    \end{tabular}
    }
 \caption{\textbf{Text-video clip retrieval} results on YouCook2 and MSR-VTT. ``FT'' denotes fine-tuning on the training set.}
      \label{tab:retrieval_youcook}
\end{table}

\subsubsection{Action segmentation}
The action segmentation task in COIN is to design an action label for a video at the frame-level.
To apply \ours to action segmentation, we fine-tune \ours by adding a linear classifier upon the output features for dense frame labeling. 
We do not feed the text descriptions during the fine-tuning process.
The results are shown in Table~\ref{tab:action_segmentation}. The baseline methods are conducted by~\cite{tang2019coin}. 
Notably, \ours significantly outperforms the baselines with more than 20\% improvements. It shows that the pre-trained \ours can deal with only visual inputs when linguistic descriptions are absent. When we remove the regional information, we observe a performance drop compared to our full model. It shows that  detailed local cues are important to the dense frame labeling task.

\subsubsection{Action step localization}
\label{sec:action_step_localization}
We evaluate action step localization on CrossTask.
To fairly compare to \cite{miech2019howto100m}, we do not fine-tune on the target dataset.
We regard the step action label as the text description and directly feed the text-video pair to \ours.
We regard the prediction for the first token ``[CLS]'' as the relevance score of this clip belonging to the label. We choose the action with the max relevance score as the final prediction.
The results are shown in Table~\ref{table:action_step_localization}.
\ours significantly outperforms TVJE~\cite{miech2019howto100m} with a large margin, \ie, the average improvement is 7\%. We achieve even better than the supervised baseline. We remove the region cues to have a fair comparison to~\cite{miech2019howto100m}, as \cite{miech2019howto100m} does not use object detection features for video and text matching. The results of ``\ours w/o region cues'' also substantially outperform \cite{miech2019howto100m}, demonstrating the effectiveness of \ours pre-training. Our full \ours model further improves performance by 4\%. This validates that regional information is an important source that provides detailed  local object features for text-and-video matching.

\subsubsection{Text-video clip retrieval}
We evaluate \ours on the task of video clip retrieval with natural language queries. 
Given a linguistic query, it aims to rank the video clips from a gallery video set.
We use the following metrics for evaluation~\cite{miech2019howto100m}, \ie, Recall@1 (R@1), Recall@5 (R@5), Recall@10 (R@10) and the median rank (Median R).
We evaluate \ours on YouCook2 and MSR-VTT. We followed \cite{miech2019howto100m} to conduct the YouCook2 evaluation.
The results are shown in Table~\ref{tab:retrieval_youcook}.
\ours significantly outperforms TVJE \cite{miech2019howto100m} and other baselines. TVJE trains a  ranking loss on the HowTo100M dataset. It shows \ours is a better pre-training framework for video-text joint representation learning. Notably, our pre-trained model achieves better retrieval performance than the finetuned TVJE model (``TVJE +FT'') on YouCook2. It shows the superiority of \ours in self-supervised video-text representation learning.
In MSR-VTT, \ours outperforms TVJE by 1.1\% on R@1 when no labeled data is accessed. Note that JSFusion \cite{yu2018joint} is a supervised method that leverages labeled video and text pairs for training.

\begin{table}[tb]
\centering
\small
\begin{tabular}{l|cc|}
\hline
Method  & {\footnotesize Accuracy}   \\ 
\midrule
LSTM-fusion \cite{yu2018joint}             & 38.3        \\ 
C+LSTM+SA-FC7 \cite{torabi2016learning}     & 60.2      \\
VSE-LSTM \cite{kiros2014unifying}            & 67.3       \\ 
SNUVL  \cite{yu2016video}                & 65.4      \\ 
EITanque \cite{kaufman2017temporal}          & 65.5       \\
CT-SAN           \cite{yu2017end}       & 66.4     \\ 
MLB \cite{kim2016hadamard}                   & 76.1       \\
JSFusion  \cite{yu2018joint}                                  & 83.4  \\ 
\midrule
\ours & \textbf{85.7} \\
\bottomrule
\hline
\end{tabular}
\caption{
\textbf{Video question answering (multiple-choices)} results on MSR-VTT.
}
\label{tbl:results_mc}
\end{table}

\begin{table}
\centering
\small
\begin{tabular}{lc }
\toprule
Method  & {\footnotesize Accuracy}   \\ 
\midrule
Text-only BLSTM \cite{maharaj2017dataset} & 32.0   \\
Text-only Human \cite{maharaj2017dataset} & 30.2  \\
GoogleNet-2D + C3D \cite{maharaj2017dataset}  & 35.7                     \\
Merging-LSTM \cite{mazaheri2016video}   & 34.2                     \\
SNUVL               \cite{yu2016video}  & 38.0                     \\ 
CT-SAN            \cite{yu2017end}     & 41.9                     \\
LR/RL LSTMs            \cite{mazaheri2017video} & 40.9       \\ 
JSFusion \cite{yu2018joint}                & {45.5}                    \\
\midrule
\ours & \textbf{48.6} \\
\bottomrule
\end{tabular}
\caption{\textbf{Video question answering (fill-in-the-blank)} results on LMSDC.
}
\label{tbl:results_mcfib}
\end{table}

\subsubsection{Video question answering.}
We evaluate \ours on VideoQA tasks.
For multi-choice VideoQA, we fine-tune the pre-trained \ours on the MSR-VTT training set. The video-text pairs are fed to \ours. We use a linear classifier upon the output feature.
We use a small learning rate of 0.0001 and use Adam optimizer for training. At  the inference time, we fed each candidate with the video clip to \ours. The final choice is made by selecting the candidates with the max matching score.
The results are shown in Table~\ref{tbl:results_mc}. We compare to many baselines in this task. Without fancy joint modeling, \ours significantly outperforms JSFusion \cite{yu2018joint} by 2.3\%. It shows \ours's strong generalization from a large-scale dataset.
We additionally evaluate on another VideoQA task on LSMDC, \ie, fill-in-the-blank VideoQA.
We report the prediction accuracy on the public test set and the results are shown in Table \ref{tbl:results_mcfib}. It shows \ours is capable of learning generalizable features that it achieves considerable gains when the target video domains are movies.

\section{Conclusion}
In this paper, we introduce \ours for joint video-text modeling in a self-supervised way. We directly model both global and local visual cues for fine-grained visual and linguistic relation learning. \ours takes three sources of information as input, \ie, global actions, local regional objects, and linguistic descriptions. The novel tangled transformer further enhances the communications between the three sources.
Quantitative results on five video-text benchmarks demonstrate the effectiveness of \ours. In the future, we will consider evaluating \ours on video action recognition and detection. We will also improve \ours by designing more powerful modules for video and text modeling.

\noindent\textbf{Acknowledgements.} This work is supported by ARC DP200100938.

{\small
\bibliographystyle{ieee_fullname}
\bibliography{egbib}
}

\end{document}